\documentclass[letterpaper, 10 pt, conference]{ieeeconf}  

\IEEEoverridecommandlockouts                              

\usepackage{xcolor}
\usepackage{soul} 
\usepackage{algpseudocode}
\usepackage{algorithm}
\usepackage{amsmath}
\usepackage[mathscr]{eucal}
\usepackage{amssymb}
\usepackage{mathrsfs}
\usepackage{graphicx}
\usepackage{tikz,tkz-euclide}
\tikzset{>=latex}
\usepackage{booktabs}
\usepackage[colorinlistoftodos]{todonotes}
\usepackage{bm}
\usepackage{multirow}
\usepackage{changepage}
\usepackage{caption}
\usepackage{subfigure}

\usepackage{bm}

\usepackage{amsthm} 
\usepackage{amsfonts}
\usepackage{amsmath}

\theoremstyle{definition}

\usepackage{balance}

\definecolor{blush}{rgb}{0.87, 0.36, 0.51}

\DeclareMathOperator{\E}{\mathbb{E}}

\let\amssymbboxplus\boxplus
\let\amssymbboxminus\boxminus

\renewcommand{\boxplus}{\mathbin{\mathop\amssymbboxplus}}
\renewcommand{\boxminus}{\mathbin{\mathop\amssymbboxminus}}

\title{\LARGE \bf
Continuous Planning for Inertial-Aided Systems
}

\author{Mitchell Usayiwevu, Fouad Sukkar, Chanyeol Yoo, Robert Fitch and Teresa Vidal-Calleja
\thanks{Authors are with the UTS Robotics Institute, University of Technology Sydney, 2007, Ultimo, NSW, Australia. Corresponding author {\tt Mitchell.Usayiwevu@student.uts.edu.au.}}
\thanks{This research is supported by the UTS International Research Scholarship and the ITT Centre for Collaborative Robotics in Advanced Manufacturing funded by ARC (Project ID: IC200100001).}}

\begin{document}

\maketitle
\thispagestyle{empty}
\pagestyle{empty}

\begin{abstract}
    Inertial-aided systems require continuous motion excitation among other reasons to characterize the measurement biases that will enable accurate integration required for localization frameworks. This paper proposes the use of informative path planning to find the best trajectory for minimizing the uncertainty of IMU biases and an adaptive traces method to guide the planner towards trajectories which aid convergence. The key contribution is a novel regression method based on Gaussian Process (GP) to enforce continuity and differentiability between waypoints from a variant of the RRT$^*$ planning algorithm. We employ linear operators applied to the GP kernel function to infer not only continuous position trajectories, but also velocities and accelerations. The use of linear functionals enable velocity and acceleration constraints given by the IMU measurements to be imposed on the position GP model. The results from both simulation and real world experiments show that planning for IMU bias convergence helps minimize localization errors in state estimation frameworks.
\end{abstract}

\section {Introduction}

Localization is an important part of robotic navigation, in which the robot uses information from its onboard sensors to estimate its location over time \cite{siegwart2011introduction,panigrahi2021localization}. This field of research has grown in popularity over the past two decades, with the majority of the localization approaches being split between optimization-based~\cite{chang2016improved,leutenegger2015keyframe} or filtering-based~\cite{bloesch2015robust}. Robust frameworks usually employ more than one sensing modality for localization~\cite{cadena2016past}, with cameras, LiDARs and \emph{Inertial Measurement Units}~(IMUs) among the most commonly used sensors. The estimation algorithms used with any of these sensors need to account for sensor noises which corrupt the measurements. With IMUs however, there exists an additional layer of complexity introduced by the biases of the accelerometer and the gyroscope. These biases are embedded within the sensor measurements and need to be accounted for during state estimation in inertial-aided systems. Most approaches take a passive approach of handling the bias errors, where the estimation is carried out to find the robot location and bias with the same importance on each task. 

In this work, we propose a more active approach of performing localization, in the context of inertial-aided systems. We use informative path planning to find the best possible trajectory to estimate the IMU biases, thereby leading to more accurate localization estimates. Our framework exploits an adaptive technique to guide the planning to minimize bias uncertainties first and localization uncertainty after convergence. This adaptive technique is used as the cost function in the RRT$^*$ variant that generates a set of discrete waypoints. However, to  generate  coherent  IMU  readings to explore the space, the system requires continuous trajectories that are twice-differentiable in position and once differentiable in orientation trajectories. In order to ensure such trajectories, Gaussian Process (GP) regression is used to interpolate the waypoints coming from our sampling-based planner. Linear operators are applied to the kernel function of underlying GP on position in order to infer the first and second derivatives. Additionally, the use of linear functionals enable velocity and acceleration constraints to be added in the GP model as part of the measurement vector. 

The key contribution of this work is two-fold. 1) An informative path planning algorithm that adapts to prioritize convergence of IMU biases to improve localization accuracy. 2) A method to generate continuous and differentiable paths based on GP regression and linear operators, that allows embedding of constraints in the first and second derivatives (velocity and acceleration measurements) in the position trajectory.

\begin{figure}[t]
\centering
    \subfigure[]{
         \includegraphics[width=0.4\columnwidth]{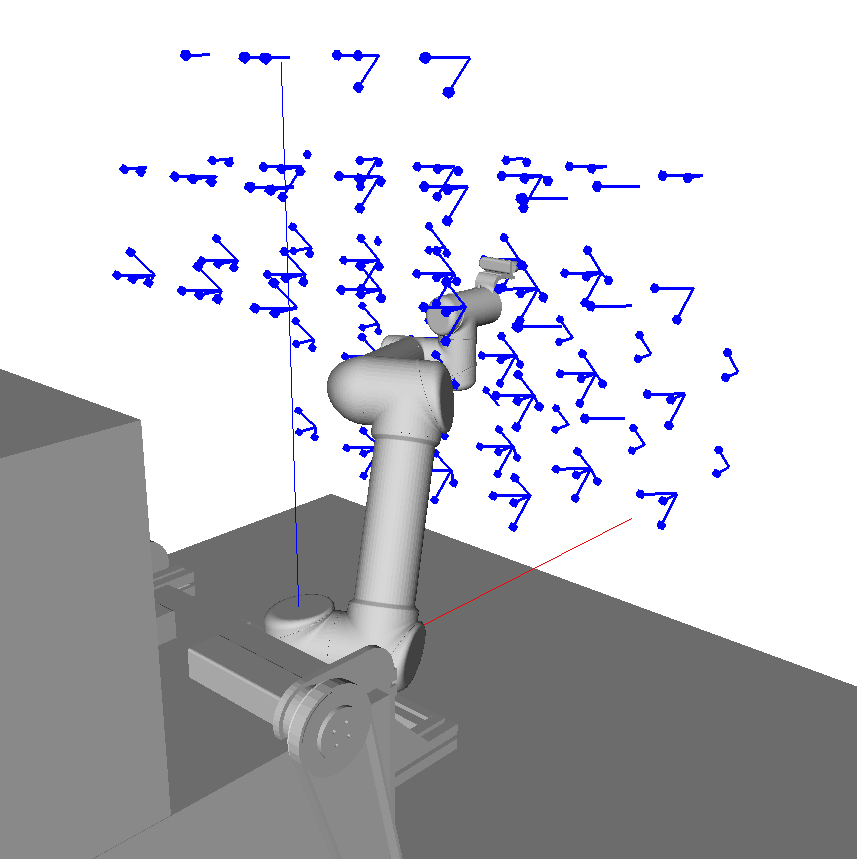}
         \label{subfig:subspace_sim}}
     \subfigure[]{
         \includegraphics[width=0.4\columnwidth]{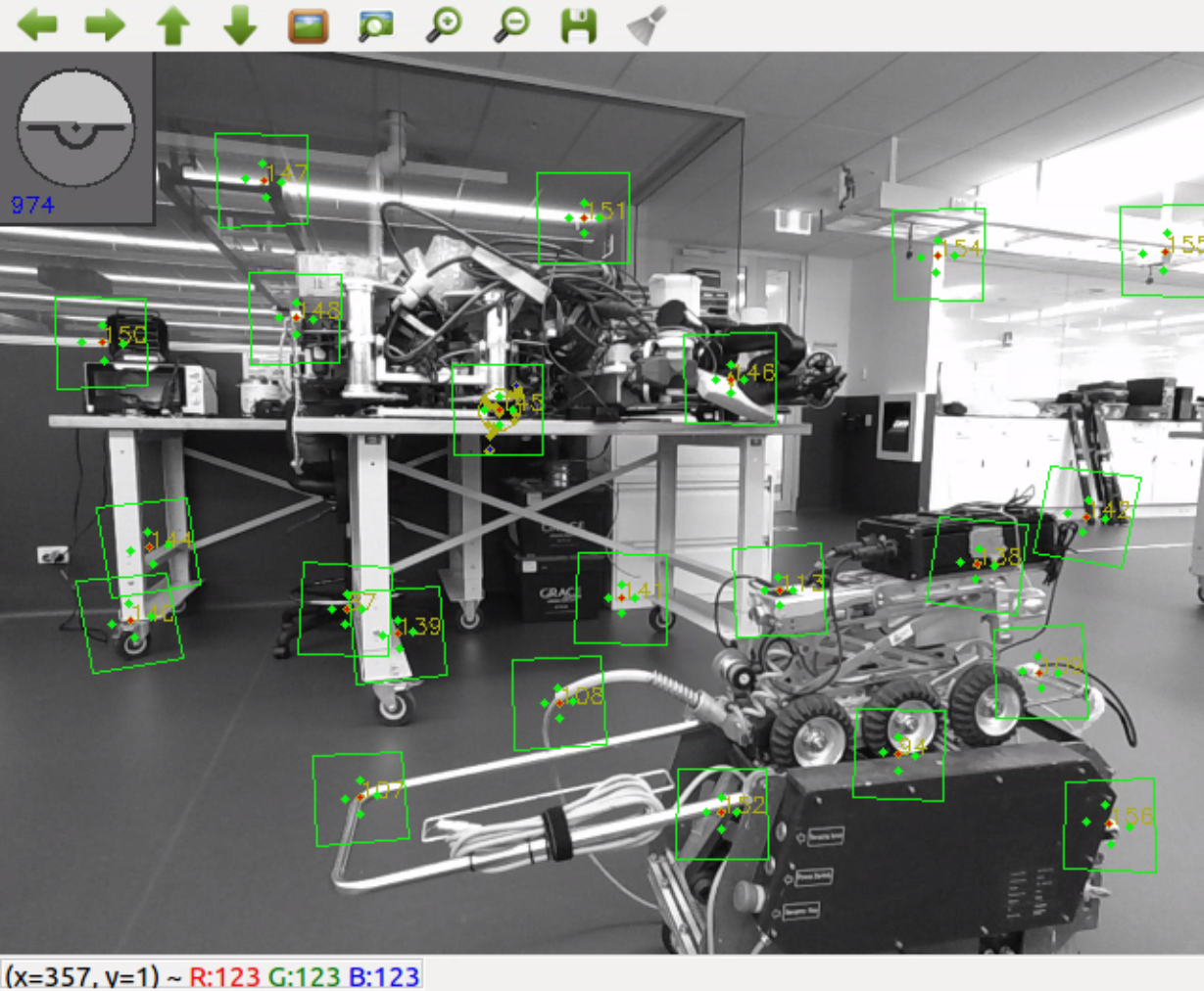}
         \label{subfig: rovio features}
         }
         
    \subfigure[]{
         \includegraphics[width=0.4\columnwidth]{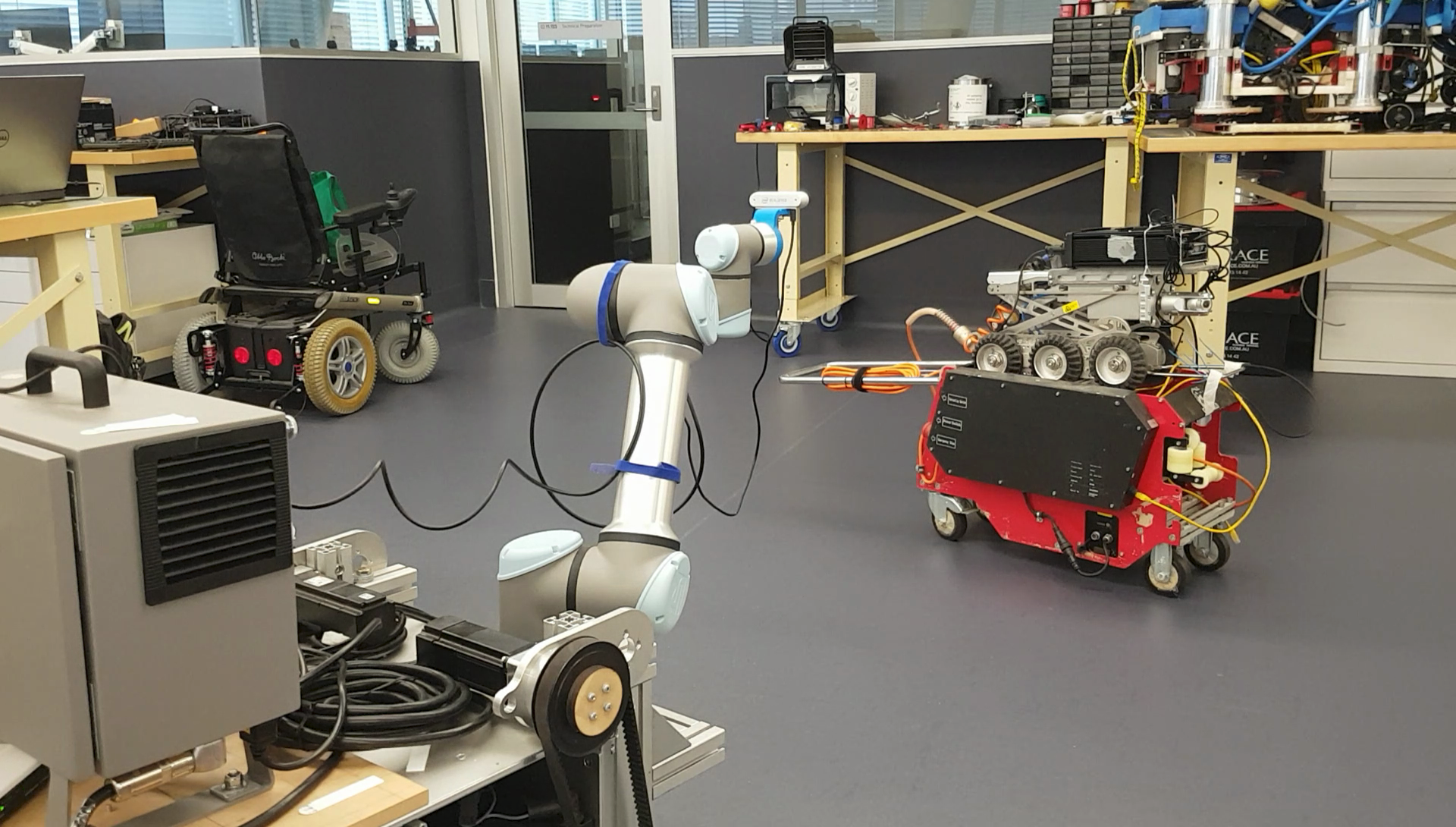}\label{subfig:real_ur5}}
     \subfigure[]{
         \includegraphics[width=0.4\columnwidth]{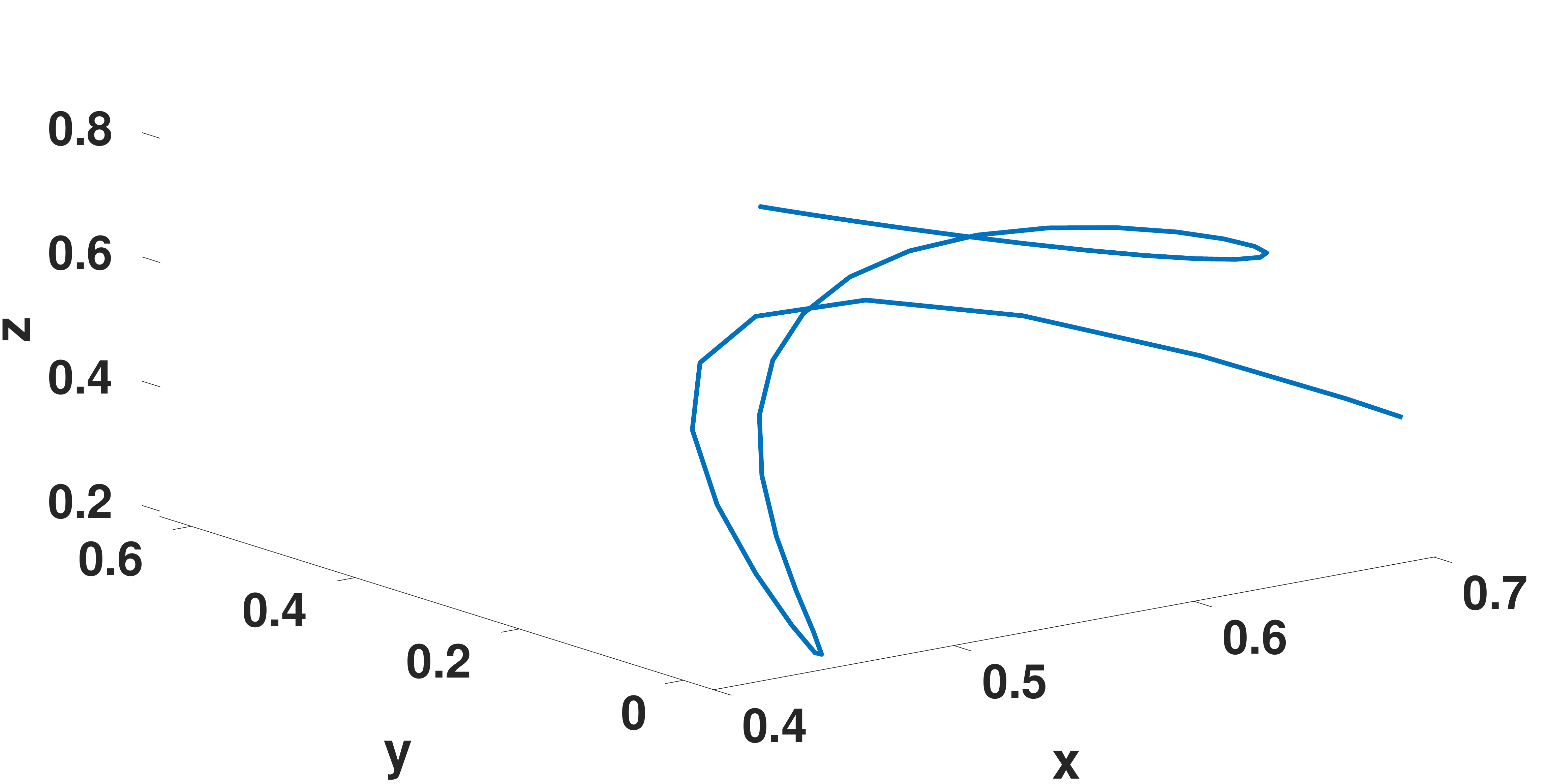}
         \label{subfig: trajectory sample}
         }
      \caption{A 6-DOF UR5 arm executing an informative trajectory from our planner that prioritizes IMU bias convergence for increased localization accuracy. \protect\subref{subfig:subspace_sim} shows the roadmap used by the planner to bias trajectories towards smooth paths, while the visual features used by the estimation algorithm are shown in~\protect\subref{subfig: rovio features}. The scene with the UR5 arm is shown in~\protect\subref{subfig:real_ur5} and~\protect\subref{subfig: trajectory sample} is the 3D executed trajectory.}
     \label{fig:fig_roadmap_robot_features_trajectory}
     \vspace{-2em}%
\end{figure}

Moreover, our algorithm is validated in a series of simulations and real-world experiments. 
The real-world experiments are carried out on a UR5 arm as shown in Fig.~\ref{fig:fig_roadmap_robot_features_trajectory}, where our planner which prioritizes IMU biases convergence by using adaptive traces outperforms the non-adaptive method. 

\section{Related Work}
There exists a number of works in the literature that have tackled continuous planning for IMU-aided systems. A common approach used, which is optimization based, generates minimum snap trajectories for quadrotor systems~\cite{mellinger2011minimum}. The approach generates trajectories that minimize the square of the norm of the fourth derivative of position (snap) using \emph{Quadratic Program}~(QP). Work in \cite{mellinger2012mixed,burke2021fast} improve the numerical stability of the underlying QP and make it possible to have long-range trajectories composed of many segments for a finite time. Our regression method based on GPs can also deal with long-range trajectories and include as an addition their associated uncertainty. 

Another common method of generating continuous trajectories is through parametric polynomial curves called Bezier curves. In~\cite{yassine2022robust}, the authors compare the control performance of continuous trajectories generated by Bezier curves and B-splines. A major drawback of using Bezier curves is that they do not consider the dynamics of the system in the interpolation. In \cite{hitz2017adaptive}, authors use B-splines for continuous trajectories within an \emph{Informative Path Planning}~(IPP) framework to find an efficient path for maximizing information collected by the robot while respecting time and energy constraints~\cite{wakulicz2022informative}. Similarly, authors in \cite{bahnemann2017sampling} and \cite{usayiwevu2020information} employ sampling-based motion planning to explore the space of possible beliefs and find a maximally informative trajectory within a user-defined budget to reduce model parameter uncertainty. Although this is similar to our work, in that we use an IPP for maximizing information to reduce localization uncertainty, we prioritize IMU bias convergence in order to improve the quality of the bias estimation and ultimately improve state estimation.

GPs are used in motion planning in \cite{mukadam2018continuous} and \cite{marchant2014bayesian}. In \cite{mukadam2018continuous}, the GP representation is tightly coupled to the gradient descent-based optimization algorithm for full state estimation. Our proposed framework uses GP regression in a loosely coupled manner that allows the planner to focus on reducing bias uncertainty. Authors in \cite{marchant2014bayesian} use GP regression for space-time modeling then apply Bayesian Optimization to estimate the best path to collect new observations in an exploratory manner. Unlike their work, we use GP regression to interpolate between sample points from our planner that finds the best trajectory to improve localization accuracy. Additionally, application of linear operators to the kernel function of underlying GP in position in our method allows inference the first and second derivatives.

Inertial based active localization, active SLAM, navigation and exploration can be found in \cite{qin2019autonomous}, \cite{elisha2017active} and \cite{papachristos2017autonomous}. These  works operate by providing approaches in which the action the robot takes and the measurements collected are the most efficient for reducing localization uncertainty. Authors of \cite{liu2005minima} improve localization accuracy for inertial-aided systems by reducing the noise level in the raw acceleration measurements. Despite the direct link between localization estimates and the IMU bias estimates, most approaches take a passive approach of handling the bias errors, where the estimation is carried out to find the robot location and bias with the same importance on each task. To the best of our knowledge, no other work exploits IMU bias convergence to guide planning for improved localization estimates. The closest work to ours is presented in \cite{elisha2017active}, were the authors propose an active calibration framework of the intrinsic parameters such as IMU bias, which allows the robot to select the best path that improve its state estimation. The main distinction between our work and theirs is that their approach is based on belief space planning, while we employ an IPP algorithm that prioritizes convergence of IMU biases to improve localization accuracy. Our novel approach employs an adaptive technique in which the biases uncertainty is used to guide the planning before convergence of the bias uncertainties and localization uncertainty is used after convergence. 

\section{Problem Statement and Overview}

\subsection{Inertial Based Systems}
Consider an inertial-aided system whose state can be estimated by any probabilistic estimation framework, and the state is modelled as a multivariate Gaussian distribution $\mathcal{N}(\mathbf{x},\,\mathbf{P})$. The mean state defined as
 $   \mathbf{x} = (\mathbf{r, v, R, }\mathbf{b}_f, \mathbf{b}_w, \mathbf{c, z}),$
where $\mathbf{r}$ is the position of the IMU in the world frame $W$, $\mathbf{v}$ is the velocity of the IMU in $W$, $\mathbf{R}\in$ $SO(3)$ is the orientation (represented as a rotation matrix from $W$ to the IMU frame $I$), $\mathbf{b}_f$ is additive bias on accelerometer, $\mathbf{b}_w$ is additive bias on gyroscope, $\mathbf{c}$ and $\mathbf{z}$ are the linear and rotational part of extrinsic parameters between the IMU and the exteroceptive sensor used, and $\mathbf{P}$ is the state covariance matrix. 

The process model and measurement model of the additional exteroceptive sensor are defined as
\begin{align}
    \mathbf{\dot{x}}(t) &= f(\mathbf{x}(t),\mathbf{u}(t), \boldsymbol{\epsilon}(t) )  \\
    \mathbf{y}(t) &= h(\mathbf{x}(t), \boldsymbol{\upsilon}(t)) ,
\end{align}
where $\mathbf{u}(t)$ is the control input, the process noise is $\boldsymbol{\epsilon}(t) \sim \mathcal{N}(0,\,\boldsymbol{\Sigma}_{\epsilon}(t))$ and the measurement noise is $\boldsymbol{\upsilon}(t) \sim \mathcal{N}(0,\,\boldsymbol{\Sigma}_{\upsilon}(t))$.

The IMU provides linear acceleration $\tilde{\mathbf{f}}(\mathbf{t}_{i})$ and angular velocity measurements $\tilde{\boldsymbol{\omega}}(\mathbf{t}_{i})$ at time $\mathbf{t}_{i}$ with $i = (1,....,T)$ in the inertial reference frame. The linear acceleration of the IMU in $W$ is denoted as $\mathbf{f}_{W}$, while $\boldsymbol{\omega}$ is the angular velocity of the IMU frame relative to $W$. The relationship between the IMU measurements and $\mathbf{f}_{W}(\mathbf{t}_{i})$ and $\boldsymbol{\omega}(\mathbf{t}_{i})$ is given by,
\begin{align}
    \tilde{\mathbf{f}}(t) &= \mathbf{R}_{W}^{t}(t)^{\top}(\mathbf{f}_{W}(t)- \mathbf{g}) + \mathbf{b}_{f}(t) + \boldsymbol{\eta_{f}}(t) \\
    \tilde{\boldsymbol{\omega}}(t) &= \boldsymbol{\omega}(t) + \mathbf{b}_{\omega}(t) + \boldsymbol{\eta_{\omega}}(t)\,,
\end{align}
where $\mathbf{g}$ is the gravity vector in $W$, and $\boldsymbol{\eta}_{f}$ and $\boldsymbol{\eta}_{\omega}$ are zero-mean Gaussian sensor noises with
covariance matrix $\Sigma_{\eta_{f}}$ and $\Sigma_{\eta_{\omega}}$ for the linear accelerations and angular velocities respectively.

At time $t$ given an IMU, the system kinematics $f(\mathbf{x}(t),\mathbf{u}(t), \boldsymbol{\epsilon}(t))$ is given by:
\begin{align}
    \dot{\mathbf{R}}_{W}^{t}(t) &= \mathbf{R}_{W}^{t}(t) (\tilde{\boldsymbol{\omega}}(t) - \mathbf{b}_{\omega}(t) - \boldsymbol{\eta}_{\omega}(t))^{\wedge}\\
    \dot{\mathbf{v}}_{W}^{t}(t) &=  \mathbf{R}_{W}^{t}(t) (\tilde{\mathbf{f}}(t) - \mathbf{b}_{f}(t) - \boldsymbol{\eta}_{f}(t)) + \mathbf{g}  \\
    \dot{\mathbf{r}}_{W}^{t}(t) &= \mathbf{v}_{W}^{t}(t)\,, 
\end{align}
and the IMU sensor biases modelled by a Brownian motion,
\begin{align}
    \dot{\mathbf{b}}_{f}(t) &= \boldsymbol{\eta}_{\mathbf{b}_{f}}(t) \\
    \dot{\mathbf{b}}_{\omega}(t) &= \boldsymbol{\eta}_{\mathbf{b}_{\omega}}(t)\,,
\end{align}
where $\boldsymbol{\eta}_{\mathbf{b}_{f}}$ and $\boldsymbol{\eta}_{\mathbf{b}_{\omega}}$ are zero-mean Gaussian noise of the accelerometer and gyroscope biases, with variances given by $\Sigma_{b_f}$ and $\Sigma_{b_{\omega}}$ respectively.

Thus, the control input is given by,
\begin{align}
     \mathbf{u}(t) = 
     \begin{bmatrix}
          \tilde{\mathbf{f}}(t) - \boldsymbol{\eta_{f}}(t) \\
          \tilde{\boldsymbol{\omega}}(t) - \boldsymbol{\eta_{\omega}}(t)
     \end{bmatrix}
     \quad
     \boldsymbol{\epsilon}(t) =
     \begin{bmatrix}
        \boldsymbol{\eta}_{\mathbf{b}_{f}}(t) \\
        \boldsymbol{\eta}_{\mathbf{b}_{\omega}}(t) \\
     \end{bmatrix}
     .
\end{align}
Note the symbol $^\wedge$ is the skew-symmetric matrix operator that transforms a $3\times1$ vector to a $3\times3$ matrix as
\begin{align}
    \boldsymbol{\omega}^\wedge =     
    \begin{bmatrix}
         \omega_{1}\\
         \omega_{2}\\
         \omega_{3}\\
    \end{bmatrix}
    ^\wedge
    =   
      \begin{bmatrix}
         0&-\omega_{3}&\omega_{2}\\
         \omega_{3}&0&-\omega_{1}\\
         -\omega_{2}&\omega_{1}&0\\
    \end{bmatrix}\,.
\end{align}

\subsection{Problem formulation}
\label{sec:problem formulation}
Given an inertial-aided system and an associated estimation framework moving in an unknown environment, the aim is to find the continuous optimal trajectory $\boldsymbol{\pi}^{*}$ of the system, in the space of all trajectories $\psi$ for maximum gain in the information-theoretic measure,
    \begin{align}\label{eq:ipp}
        \boldsymbol{\pi^{*}} &= \underset{\pi \in \psi}{\text{argmax}} \;
             \frac{\text{I}[\text{M}(\pi)]}{\text{T}(\pi)},\\
             \text{s.t.}
            &\ C(\pi) \leq B \nonumber,
    \end{align}
    where~$\text{I}[\cdot]$ is the utility function that evaluates the information gain in localization. The
    function $\text{M}(\cdot)$ obtains discrete sensor measurements along the trajectory $\boldsymbol{\pi}$ with $\text{T}(\cdot)$ as corresponding travel time. The cost of the path $\text{C}(\cdot)$ given by the planner cannot exceed a predefined budget $\text{B}$. The utility function above is formulated to compute the expected reduction in IMU biases uncertainty and robot localization uncertainty.

\subsection{Overview}
 
We propose an Informative Path Planning framework as described in Section~\ref{sec:problem formulation} that directly takes into account the impact of the biases $\mathbf{b}_f$ and $\mathbf{b}_\omega$ embedded in the IMU measurements $\tilde{\mathbf{f}}$ and $\tilde{\boldsymbol{\omega}}$ to maximize localization information gain or in other words minimize localization uncertainty. Given a trajectory, a state estimation framework is used to generate a prior map of the environment~$\mathcal{E}$ and to produce initial estimates of the state and its associated covariance. The last state of the prior trajectory is set as the start node for the planning algorithm. An RRT$^{*}$ planner is used to build a decision tree by sampling in the linear position and orientation space. GP regression is then used to connect in-between two nodes and propagate uncertainties to evaluate the proposed IPP metric. The planner considers poses that have the most \emph{excitation} in the acceleration and angular velocity space which helps the IMU biases to converge quicker and producing more accurate localization estimates. Note that our planner can work with any filtered-based inertial-aided estimation framework as we shown in the experiments section using two existing frameworks.

\section{GPs for Continuous Trajectories} \label{sec:GP}
A \emph{Gaussian Process} (GP) is a collection of random variables, any finite number of which have a joint Gaussian distribution \cite{rasmussen2003gaussian}. It is completely specified by its mean function~$\mu(\boldsymbol{t})$ and covariance function~$k(\boldsymbol{t,t'})$ for a real function~$\xi: \mathbb{R}^d \mapsto \mathbb{R}^{s}$ 
%
 \begin{align}
     \mu(\mathbf{t}) &= \E[\xi(\mathbf{t})] \\
     k(\mathbf{t,t'}) &= \E[(\xi(\mathbf{t}) - \mu(\mathbf{t}))(\xi(\mathbf{t'}) - \mu(\mathbf{t'}))]
     .
 \end{align}
In this work, we consider a zero-mean GP defined over time, which is used to generate continuous linear position, velocity and acceleration trajectories, angular positions and velocities used in a state estimation framework. Our GP is defined as,
\begin{align}
    \xi(\mathbf{t}) &\sim \mathcal{GP}(0 , k(\mathbf{t,t'})) \\
    \boldsymbol{\gamma}_{i} &= \xi(\mathbf{t}_{i})+ \mathbf{e}_{i}
    ,
\end{align}
where~$\boldsymbol{\gamma}_{i}, \mathbf{e}_{i} \in \mathbb{R}^{s}$ and the joint covariance of errors $\mathbf{e} =\mathbf{(e_1,e_2,...,e_n)}$ is assumed to be given by the matrix $\Sigma_{\mathbf{e}}$. 
Given a sequence of waypoints in position $\boldsymbol{\gamma =(\gamma_{1}, \gamma_{2},...,\gamma_{n})}$, the joint distribution of the observed position waypoints and the continuous position values at the test locations can be written as,
\begin{align}
    \begin{bmatrix}
         \boldsymbol{\gamma}\\ 
         \boldsymbol{\xi}_*
    \end{bmatrix}
     \sim 
     \mathcal{N} 
    \begin{pmatrix}
          \mathbf{0},
          \begin{bmatrix}
             K(\mathbf{t,t}) + \Sigma_{\mathbf{e}} & K(\mathbf{t,t}_{*}) \\
             K(\mathbf{t_{*},t}) & K(\mathbf{t_{*},t_{*}})
        \end{bmatrix}
    \end{pmatrix}
    .
\end{align}
The position posterior mean and covariance are given by the predictive Gaussian process regression as,
\begin{align}
    \bar{\boldsymbol{\xi}}_* &= \E[\boldsymbol{\xi_{*}}|\mathbf{t},\boldsymbol{\gamma},\mathbf{t}_*] = K(\mathbf{t_*,t})[K(\mathbf{t,t}) + \Sigma_{\mathbf{e}} ]^{-1}\boldsymbol{\gamma} \nonumber \\
    \text{cov}(\boldsymbol{\xi}_{*}) &= K(\mathbf{t_*,t_*}) -  K(\mathbf{t_*,t})[K(\mathbf{t,t})+ \sigma^2 _n I ]^{-1}K(\mathbf{t,t}_*)
    .\nonumber
\end{align} 
With $\mathbf{t} =
\begin{bmatrix}
         t_{1}   & t_{2} & \dots  & t_{n}\\
\end{bmatrix}^{\top}$,
\begin{align}
K(\mathbf{t_{*},t}) &= 
\begin{bmatrix}
         k(t_{1}, t_{1})  & k(t_{1},t_{2}) & \dots  & k(t_{1},t_{n})
\end{bmatrix}, \\
K(\mathbf{t,t_{*}}) &= K(\mathbf{t_{*},t})^{\top}
\end{align}
and
\begin{align}
    K(\mathbf{t,t})  = 
    \begin{bmatrix}
         k(t_{1}, t_{1})  & k(t_{1},t_{2}) & \dots  & k(t_{1},t_{n}) \\
         k(t_{2}, t_{1})  & k(t_{2},t_{2}) & \dots  & k(t_{2},t_{n})\\ 
         \vdots           &    \vdots      & \ddots & \vdots\\
         k(t_{n},t_{1})   & k(t_{n},t_{2}) & \dots  & k(t_{n},t_{n}) 
    \end{bmatrix}.
\end{align}
Suppose we want to use this model for inferring velocities and acceleration. GPs are adept at predicting not only the posterior mean and covariances of the function values but their derivatives as well \cite{sarkka2011linear}. This is because differentiation is a linear operator on the space of functions and hence the derivative of a GP is another GP. So velocity and acceleration functions obtained from applying linear operators to the position function are GPs as well. We choose to use the Square Exponential (SE) kernel given that it is analytically infinitely differentiable. 

Consider the linear operator $\mathcal{L}^t$ applied on the function $\boldsymbol{\xi(t)}$ as follows,
\begin{align} 
    \boldsymbol{\phi(t)} &= \mathcal{L}^t_{\phi}\boldsymbol{\xi(t)}\\
     \boldsymbol{\zeta(t)} &= \mathcal{L}^t_{\zeta}\boldsymbol{\phi(t)} = \mathcal{L}^t_{\zeta}\mathcal{L}^t_{\phi}\boldsymbol{\xi(t)} \label{eq:operator}
     ,
\end{align}
where $\mathcal{L}^{t}$ = $\mathbf{d(t)}$ is the derivative operator.

Note that the linear operators~$\mathcal{L}^{t}$ are not matrix multiplication, but can be thought of as operators acting on a function and return another function with the same input domain as the input function $\phi: \mathbb{R}^d \mapsto \mathbb{R}^{s}$. When the operator is applied twice on the kernel (e.g., $\mathcal{L}^t_{\zeta}\mathcal{L}^t_{\phi}\boldsymbol{\xi(t)}$ in \eqref{eq:operator}), it is analogous of taking the partial derivative of $\boldsymbol{\xi(t)}$ with respect to $\mathbf{t}$ twice. Consequently, $\boldsymbol{\phi(t)}$ and $\boldsymbol{\zeta(t)}$ are GPs of the first and second derivative functions respectively. Linear operators can be applied on the right-hand side of the kernel function like so $\mathbf{\mathcal{L}}_{\phi}^{t_{*}} K(\mathbf{t_*,t})\mathbf{\mathcal{L}}_{\phi}^{t}$, and it is not synonymous to right multiplication by a matrix in linear algebra. The right multiplication reflects the operator is operating on the second argument of the kernel function.  

Given a training data including waypoints~$\boldsymbol{\gamma}$, velocity~$\mathbf{\dot{\boldsymbol{\gamma}}}$, and acceleration $\mathbf{\ddot{\boldsymbol{\gamma}}}$, a linear functional can be applied to the kernel matrix to incorporate derivative observations as,
\begin{align}
    \boldsymbol{\gamma}_{i} &= \xi(\mathbf{t})+ \mathbf{e}_{i} \\
    \dot{\boldsymbol{\gamma}_{i}} &= \mathbf{\mathcal{H}}_{\gamma}^t \xi(\mathbf{t})+ \mathbf{e}_{i} \\
      \ddot{\boldsymbol{\gamma}_{i}} &= \mathbf{\mathcal{H}}_{\gamma}^t \mathbf{\mathcal{H}}_{\gamma}^t \xi(\mathbf{t})+ \mathbf{e}_{i}
    ,
\end{align}
where $\mathcal{H}$ is deterministic linear functional for estimating the linear operator transformation of the signal $\mathbf{d(t)}$. Linear functionals are similar to linear operators, but they output vectors or matrices instead of functions.

\begin{figure*}
     \centering
     \subfigure[Positions over time]{
         \includegraphics[width=0.65\columnwidth]{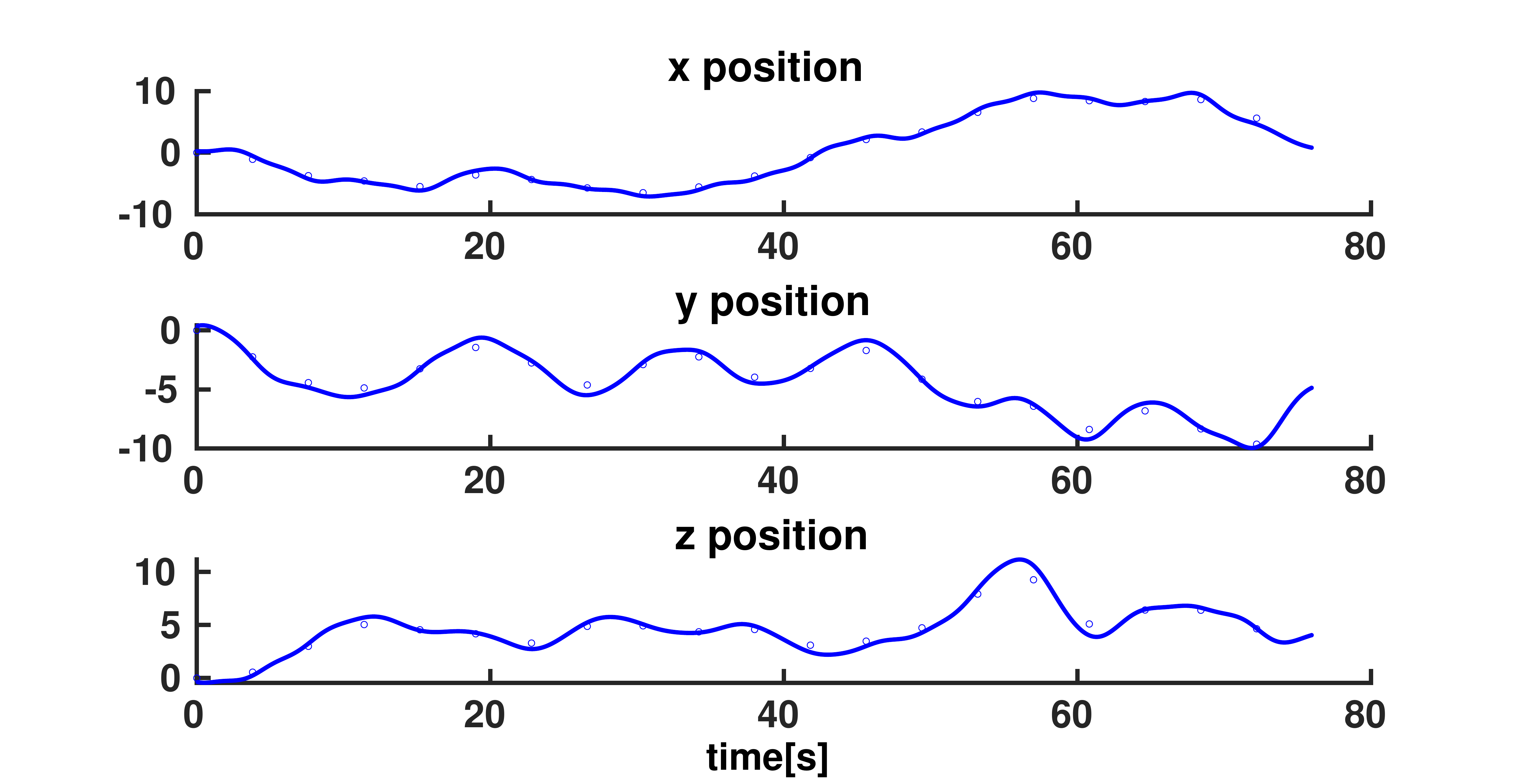}}
     \subfigure[Velocities over time]{
         \includegraphics[width=0.65\columnwidth]{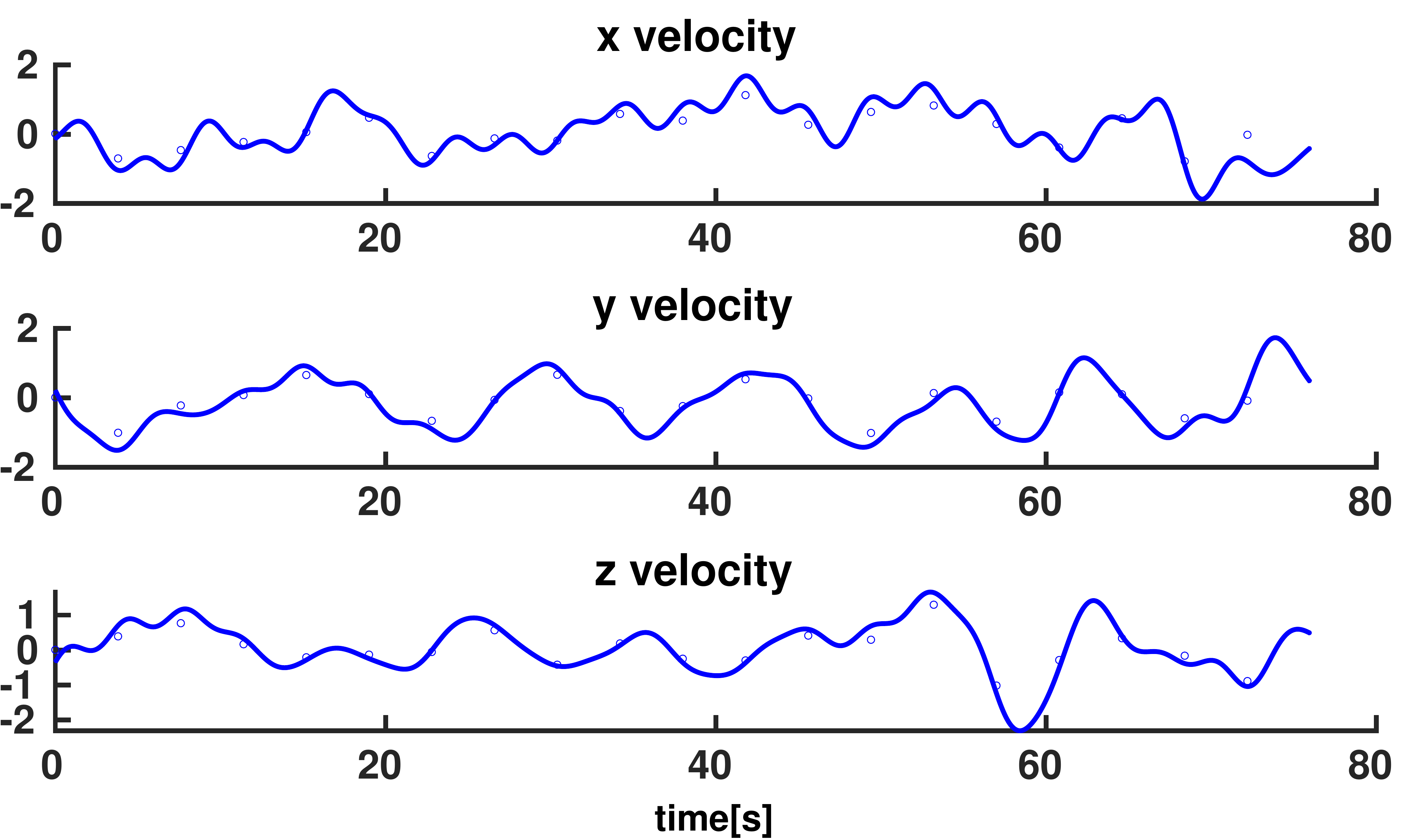}}
     \subfigure[Accelerations over time]{
         \includegraphics[width=0.65\columnwidth]{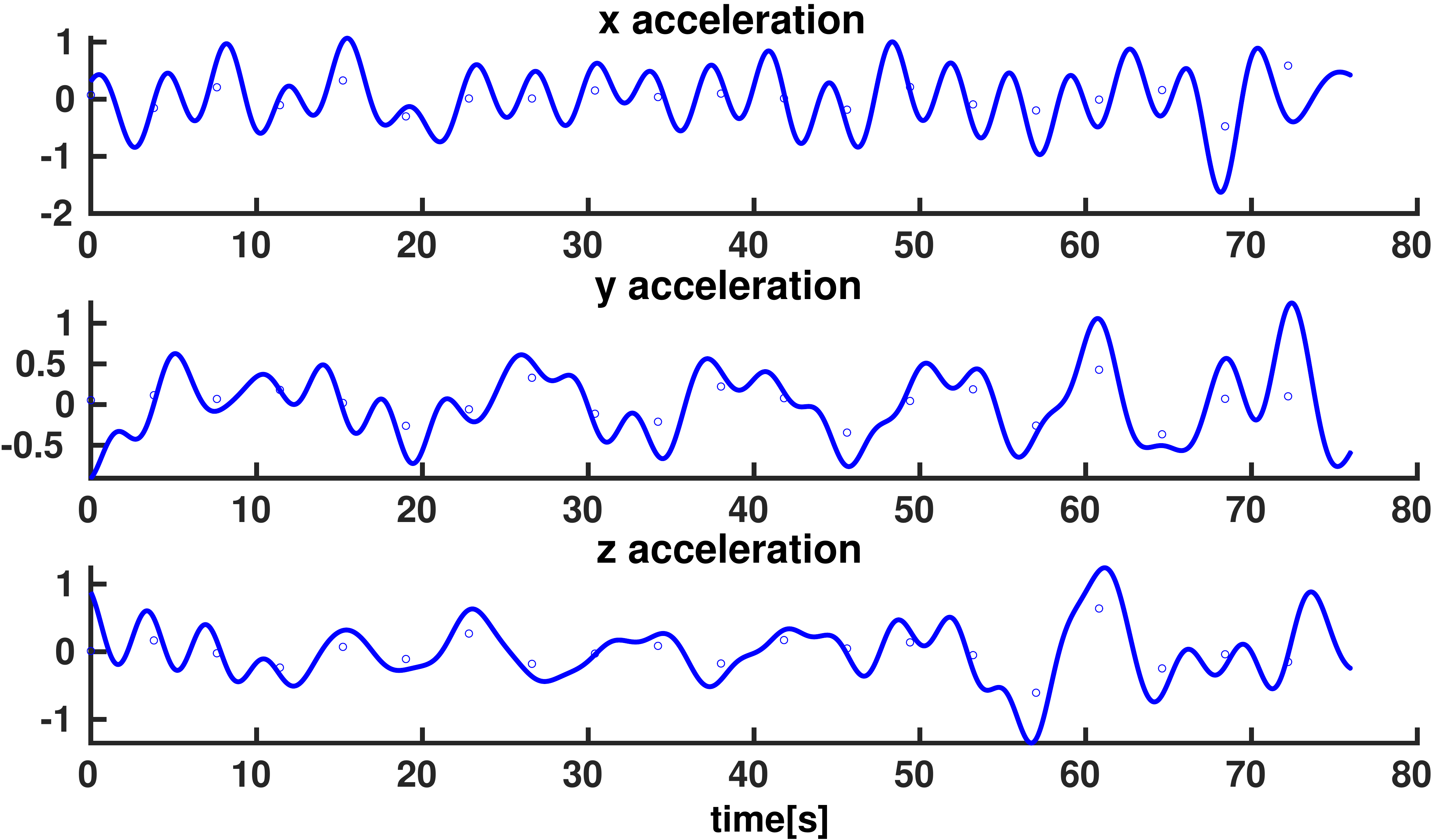}}
        \caption{Example of continuous position, velocity and acceleration trajectories in the~$(x, y, z)$ axes from the GP interpolation.}
        \label{fig:pos_vel_acc}
        \vspace{-1em}%
\end{figure*}

Through the application of a combination of linear operators and functionals to the kernel function of the underlying position GP function, we can conduct inference in the velocity and acceleration space (see Fig.~\ref{fig:pos_vel_acc}). Constraints in the velocity and acceleration space, which enforce continuity at the start and end of each segment in these spaces can also be included in the measurement (waypoints) vector and multiplied properly with the kernel function through linear functionals. The inference of $\boldsymbol{\bar{\xi}_{*},\bar{\phi}_{*}}$ and $\boldsymbol{\bar{\zeta}_{*}}$ with measurements in the position, velocity and acceleration spaces is given by,
\begin{align}\label{eq:gp_mat}
    \begin{bmatrix}
         \bar{\boldsymbol{\xi_{*}}}\\
         \bar{\boldsymbol{\phi_{*}}}\\
         \bar{\boldsymbol{\zeta_{*}}}\\
    \end{bmatrix}
    =   
      \begin{bmatrix}
         m_{1,1}&m_{1,2}&m_{1,3}\\
         m_{2,1}&m_{2,2}&m_{2,3}\\
         m_{3,1}&m_{3,2}&m_{3,3}\\
    \end{bmatrix}
    *
    \begin{bmatrix}
         \boldsymbol{\gamma}\\
        \dot{\boldsymbol{\gamma}}\\
        \ddot{\boldsymbol{\gamma}}\\
    \end{bmatrix}
    .
\end{align}
Each of the terms $m_{i,j}$ in the matrix above are defined by applying the linear operator on the GP kernel in the position space, to infer both linear and angular positions, velocities and accelerations.

Position inference;
\begin{align}
     m_{1,1} &= K(\mathbf{t_*,t})[K(\mathbf{t,t}) + \Sigma_{\mathbf{e}}]^{-1}  \nonumber \\
     m_{1,2} &= K(\mathbf{t_*,t}) \mathbf{\mathcal{H}}_{\gamma}^t [\mathbf{\mathcal{H}}_{\gamma}^t K(\mathbf{t,t})\mathbf{\mathcal{H}}_{\gamma}^t + \Sigma_{\mathbf{e}}]^{-1}  \nonumber \\
     m_{1,3} &= K(\mathbf{t_*,t}) \mathbf{\mathcal{H}}_{\gamma}^t\mathbf{\mathcal{H}}_{\gamma}^t [\mathbf{\mathcal{H}}_{\gamma}^t\mathbf{\mathcal{H}}_{\gamma}^t K(\mathbf{t,t}) \mathbf{\mathcal{H}}_{\gamma}^t\mathbf{\mathcal{H}}_{\gamma}^t + \Sigma_{\mathbf{e}}] ^{-1}\nonumber
     ,
\end{align}
Velocity inference;
\begin{align}
     m_{2,1} &= \mathbf{\mathcal{L}}_{\phi}^{t_{*}} K(\mathbf{t_*,t})[K(\mathbf{t,t}) + \Sigma_{\mathbf{e}}]^{-1} \nonumber \\
     m_{2,2} &= \mathbf{\mathcal{L}}_{\phi}^{t_{*}} K(\mathbf{t_*,t})\mathbf{\mathcal{H}}_{\gamma}^t [\mathbf{\mathcal{H}}_{\gamma}^t K(\mathbf{t,t})\mathbf{\mathcal{H}}_{\gamma}^t + \Sigma_{\mathbf{e}}]^{-1}  \nonumber \\
     m_{2,3} &= \mathbf{\mathcal{L}}_{\phi}^{t_{*}} K\mathbf{(t_*,t)}\mathbf{\mathcal{H}}_{\gamma}^t\mathbf{\mathcal{H}}_{\gamma}^t[\mathbf{\mathcal{H}}_{\gamma}^t\mathbf{\mathcal{H}}_{\gamma}^tK(\mathbf{t,t})\mathbf{\mathcal{H}}_{\gamma}^t\mathbf{\mathcal{H}}_{\gamma}^t + \Sigma_{\mathbf{e}}]^{-1} \nonumber 
     ,
\end{align}
Acceleration inference;
\begin{align}
     m_{3,1} &= \mathbf{\mathcal{L}}_{\zeta}^{t_{*}}\mathbf{\mathcal{L}}_{\phi}^{t_{*}} K(\mathbf{t_*,t})[K(\mathbf{t,t}) + \Sigma_{\mathbf{e}}]^{-1}  \nonumber \\
     m_{3,2} &= \mathbf{\mathcal{L}}_{\zeta}^{t_{*}}\mathbf{\mathcal{L}}_{\phi}^{t_{*}} K(\mathbf{t_*,t})\mathbf{\mathcal{H}}_{\gamma}^t [\mathbf{\mathcal{H}}_{\gamma}^t K(\mathbf{t,t})\mathbf{\mathcal{H}}_{\gamma}^t + \Sigma_{\mathbf{e}}]^{-1}  \nonumber \\
     m_{3,3} &= \mathbf{\mathcal{L}}_{\zeta}^{t_{*}} \mathbf{\mathcal{L}}_{\phi}^{t{*}} K(\mathbf{t_*,t})\mathbf{\mathcal{H}}_{\gamma}^t\mathbf{\mathcal{H}}_{\gamma}^t[\mathbf{\mathcal{H}}_{\gamma}^t\mathbf{\mathcal{H}}_{\gamma}^tK(\mathbf{t,t})\mathbf{\mathcal{H}}_{\gamma}^t\mathbf{\mathcal{H}}_{\gamma}^t + \Sigma_{\mathbf{e}}]^{-1} \nonumber
     .
\end{align}
The covariances are given by;
\begin{align}
  \text{cov}(\boldsymbol{\xi_{*}}) &=  K(\mathbf{t_*,t_*}) -  K(\mathbf{t_*,t})[K(\mathbf{t,t}) + \Sigma_{\mathbf{e}}]^{-1} K(\mathbf{t,t_*}) \nonumber\\
  \text{cov}(\boldsymbol{\phi_{*}}) &= \mathbf{\mathcal{L}}_{\phi}^{t_{*}} K(\mathbf{t_*,t_*})\mathbf{\mathcal{L}}_{\phi}^{t_{*}}  -  \mathbf{\mathcal{L}}_{\phi}^{t_{*}} K(\mathbf{t_*,t})\mathbf{\mathcal{H}}_{\gamma}^t [\mathbf{\mathcal{H}}_{\gamma}^t K(\mathbf{t,t})\mathbf{\mathcal{H}}_{\gamma}^t   \nonumber\\
  & + \Sigma_{\mathbf{e}}]^{-1}  \mathbf{\mathcal{H}}_{\gamma}^t K(\mathbf{t,t_*})\mathbf{\mathcal{L}}_{\phi}^{t_{*}}  \nonumber\\
 \text{cov}(\boldsymbol{\zeta_{*}}) &= \mathbf{\mathcal{L}}_{\zeta}^{t_{*}} \mathbf{\mathcal{L}}_{\phi}^{t_{*}} K(\mathbf{t_*,t_*})\mathbf{\mathcal{L}}_{\zeta}^{t_{*}} \mathbf{\mathcal{L}}_{\phi}^{t_{*}} \nonumber\\
 &- \mathbf{\mathcal{L}}_{\zeta}^{t_{*}} \mathbf{\mathcal{L}}_{\phi}^{t_{*}} K(\mathbf{t_*,t})\mathbf{\mathcal{H}}_{\gamma}^t\mathbf{\mathcal{H}}_{\gamma}^t  
 [\mathbf{\mathcal{H}}_{\gamma}^t\mathbf{\mathcal{H}}_{\gamma}^tK(\mathbf{t,t})\mathbf{\mathcal{H}}_{\gamma}^t\mathbf{\mathcal{H}}_{\gamma}^t + \Sigma_{\mathbf{e}}]^{-1} \nonumber \\
 & \mathbf{\mathcal{H}}_{\gamma}^t\mathbf{\mathcal{H}}_{\gamma}^t K(\mathbf{t,t}_*)\mathbf{\mathcal{L}}_{\zeta}^{t_{*}} \mathbf{\mathcal{L}}_{\phi}^{t_{*}}
 .
\end{align}
Note that the substitution $\mathbf{t_{*} = t} $ is done for all instances of $\mathbf{t}_{*}$ after all the operations have been performed. In the equations above, $\mathbf{t}_{*}$ is used to remove ambiguity on which variable the operator is applied on.

\section{Adaptive Trace Method}
Following~\eqref{eq:ipp}, the utility function $\text{I}[\cdot]$ evaluates the information content in the new sensor measurements with respect to the localization uncertainty. Numerous criteria exist for determining optimality in experimental designs. The most common criteria in robotics are the A-optimality and D-optimality and the choice of which one to use is made based on the application. D-optimality result in a confidence region for the parameters within minimum volume and A-optimality minimizes the average variance or the expected mean square error \cite{srivastava1974comparison}. For our purpose of minimizing localization uncertainty, A-optimality is the most efficient and formally, the A-optimality for any $n\times n$ matrix $\mathbf{P}$ is given by the trace of the matrix tr($\bullet$) as,
\begin{align}
    \text{A-optimality} = \text{tr}(\mathbf{P}).
\end{align}
The utility function I[·], which is based on our proposed Adaptive trace method is then given by:

\begin{align}\label{eq:utility_fxn}
    \text{I}(\boldsymbol{\pi}_{k:k+1}) = 
    \begin{cases}
        \text{tr}(\mathbf{P}_{\mathbf{b}_{k+1}})-\text{tr}(\mathbf{P}_{\mathbf{b}_{k}}) & \text{if} \quad \delta \geq \lambda, \\
        \text{tr}(\mathbf{P}_{\mathbf{r}_{k+1}}) -\text{tr}(\mathbf{P}_{\mathbf{r}_{k}}) & \text{otherwise}
        \end{cases}
\end{align}
where~$\mathbf{P_{\mathbf{b}}} =  [\mathbf{P_{\mathbf{b_f}}, P_{\mathbf{b_{\omega}}} ] }$ is the biases covariance matrix, $\mathbf{P_r}$ is IMU position covariance matrix, $\delta$ is the bias uncertainty, $\lambda$ is a preset threshold to determine bias uncertainty convergence, and $\pi_{k:k+1}$ is the trajectory from which evaluation measurements are obtained. 

\section{Path Planning}\label{sub_sec:planning}

To find the trajectory that minimizes localization uncertainty in the estimated state, we define the cost function between two points as the trace of either the biases $\mathbf{b}_f$ and $\mathbf{b}_w$, or the IMU position $\mathbf{r}$. The planner aims to excite the system in such a way that it prioritizes convergence of bias errors first then proceeds to focus on the trace of the IMU position. In order to achieve this, the planner alternates between two cases of minimizing either the IMU biases uncertainty or IMU position uncertainty based on whether or not the bias uncertainties have converged, as depicted in~\eqref{eq:utility_fxn}. For the planner, it can formally be defined as:
  \begin{align}\label{eq:planning_cost}
 c(\boldsymbol{\pi}_{k:k+1}) = \text{I}(\boldsymbol{\pi}_{k:k+1})\,,
\end{align}
where $c(\boldsymbol{\pi}_{k:k+1})$ is the $\mathtt{Cost}$ function associated with connecting two points connected with a trajectory $\boldsymbol{\pi}_{k:k+1}$.
 
Rapidly-exploring random tree (RRT)$^{*}$~\cite{karaman2010incremental} is the sampling-based motion planning algorithm used to generate a set of nodes that are used for evaluation in our framework. The algorithm incrementally builds a tree of feasible trajectories from an initial node $x_{init}$. At each new iteration, a new point $x_{sample}$ is sampled from the obstacle-free space $\mathcal{X}_{free}$, and connection attempts are made to vertices in $\mathcal{X}_{near}$, which is defined as vertices within a radius from $x_{sample}$. An edge is created from $x_{sample}$ to the vertex in $\mathcal{X}_{near}$ that can be connected at a minimal cost.
 
The additive cost function used for evaluating nodes is defined as:
 \begin{align}
     \mathtt{Cost}(x_{\text{sample}}) &= \mathtt{Cost(\text{Parent}}(x_{\text{sample}})) \nonumber\\
     &+  c(\mathtt{Connect} (x_{\text{sample}},x_{\text{near}}))
     .
 \end{align}  
 
After the addition of the new node $x_{sample}$ to the graph, the planner removes redundant edges from $E$, i.e, edges that are not part of a shortest path from $x_{init}$. This technique is called rewiring, and it ensures that all vertices on the tree are on a minimal cost branch. Because of the constraints the IMU modeling imposes on the system, the $\mathtt{Connect}$ function between two nodes is not a straight line as in the original RRT* algorithm. As explained above, we require smooth and continuous trajectories that are differentiable at least twice, to ensure smooth and continuous linear position, velocity and acceleration trajectories, and angular positions and velocity trajectories. We use the tailor-made interpolation method based on GP regression described in Section~\ref{sec:GP}, that guarantees that all position, velocity and acceleration trajectories from the planner meet the continuity and smoothness constraints.
 
Note that 6D sampling is carried out in position and orientation as this allows us to plan in both the Cartesian and orientation spaces, and all higher order derivatives are constrained to zero. We enforce continuity by matching the linear position, velocity, acceleration and angular position and velocity at the end of a trajectory segment with those at the start of a subsequent trajectory segment. Additionally, the trace of the covariance matrix is used as the cost function to determine the optimal trajectory which the planner returns. This trajectory is not the shortest path but rather a trajectory which leads to quicker biases convergence with better bias estimates and ultimately better accuracy in the robot localization. 

\subsection{Covariance Propagation}
For each new node sampled by our RRT-based planner, the posterior covariance matrix $\mathbf{P}_{k}^{+}$ is initially obtained by the chosen estimation framework and propagated into the future by the linearized model using the equations of the Extended Kalman filter and forecasted measurements. The trace of $\mathbf{P}_{k}^{+}$ is then used for decision making by our planner. The simulated inertial measurements are used for propagation in the prediction step, while the simulated measurements from the exteroceptive sensor are taken into account during the update step. The prediction step of the filter estimates the a-priori covariance $\mathbf{P}_{k}^{-}$, from the a-posteriori covariance estimate from the previous time step $\mathbf{P}_{k-1}^{+}$:
\begin{align}\label{eq:pred1}
     \mathbf{P}_{k}^{-} &= \mathbf{F}_{k-1}\mathbf{P}_{k-1}^{+}\mathbf{F}_{k-1}^{\top} + \mathbf{G}_{k-1}\boldsymbol{\Sigma}_{\boldsymbol{\epsilon}_{k-1}}\mathbf{G}_{k-1}^{\top}
    .
\end{align}
The covariance matrix is updated according to:
\begin{align} \label{eq:upd2}
     \mathbf{P}_{k}^{+} = (\mathbf{I}-\mathbf{K}_{k}\mathbf{H}_{k}) \mathbf{P}_{k}^{-}, 
\end{align}
where the jacobians are given by
\begin{align}
    \mathbf{F}_{k-1} &= \frac{\partial f}{\partial \mathbf{x}_{k-1}}(\mathbf{x}_{k-1}^{+}, \mathbf{u}_{k-1}, \boldsymbol{\epsilon}_{k-1}), \nonumber \\
    \mathbf{G}_{k-1} &= \frac{\partial f}{\partial \boldsymbol{\epsilon}_{k-1}}(\mathbf{x}_{k-1}^{+}, \mathbf{u}_{k-1}, \boldsymbol{\epsilon}_{k-1}), 
    \mathbf{H}_{k}   &= \frac{\partial h}{\partial \mathbf{x}_{k}} (\mathbf{x}_{k})
\end{align}
and $\mathbf{K}$ is the Kalman filter gain.

\section {Results}
We validate our approach using both simulated and hardware demonstrations.
The state estimation framework used in the simulation experiments is based on an Error State Kalman Filter (ESKF) \cite{sola2017quaternion} while ROVIO~\cite{bloesch2015robust}, which is an Iterated Extended Kalman Filter framework is used for the real-world experiments. The simulated acceleration, angular velocities and range measurements have realistic sensor noises added to them, ${0.0196}\,\mathrm{m.s^{-2}}$ and ${0.0017}\,\mathrm{rad.s^{-1}}$ for the accelerometer and gyroscope  and ${0.02}\,\mathrm{m}$ for the range measurements.  We evaluate the performance of the proposed Adaptive trace method with a greedy planner and with the variant of RRT$^{*}$ we propose as well. We also compare trajectories made with our GP regression vs the minimum snap interpolation algorithm.

\subsection{ Evaluation of proposed adaptive trace method}

We evaluate the results we get from an adaptive approach with the traces used for choosing waypoints to add to our trajectory. At each timestep, the planner picks the waypoint with the smallest trace out of the five sampled points. The adaptive approach uses the trace of the bias estimate covariance until the bias uncertainties have converged. Beyond this point, the trace of the robot position estimate covariance is used for planning. 

\begin{figure}[t]
\centering
\includegraphics[width=\linewidth]{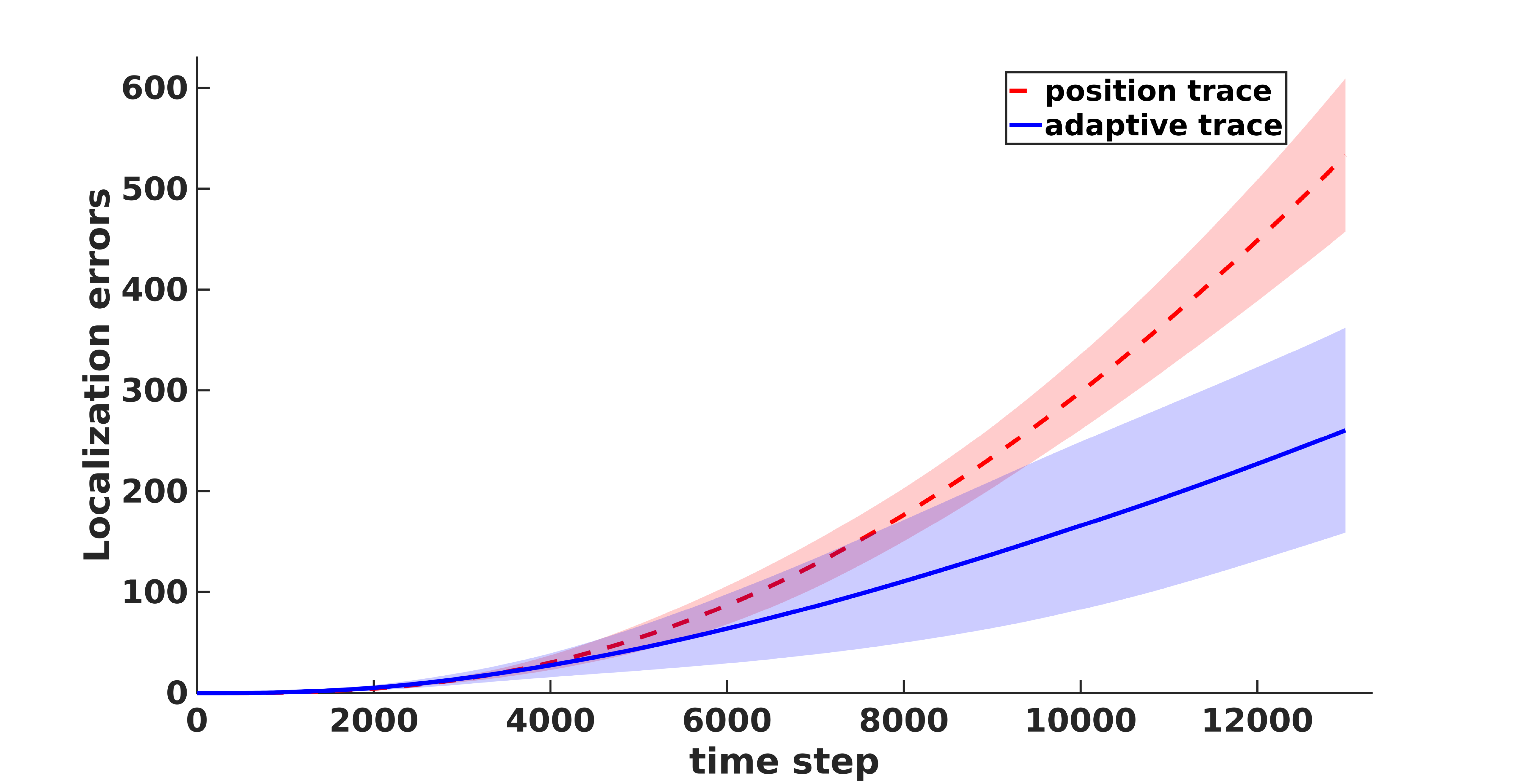}
\caption{Localization error averaged over 50 Monte-Carlo simulation for each of the 2 approaches.}
\label{fig:robot_pose_error_bounds}
\vspace{-1em}%
\end{figure}

\begin{figure}[t]
\centering
\includegraphics[width=0.9\linewidth]{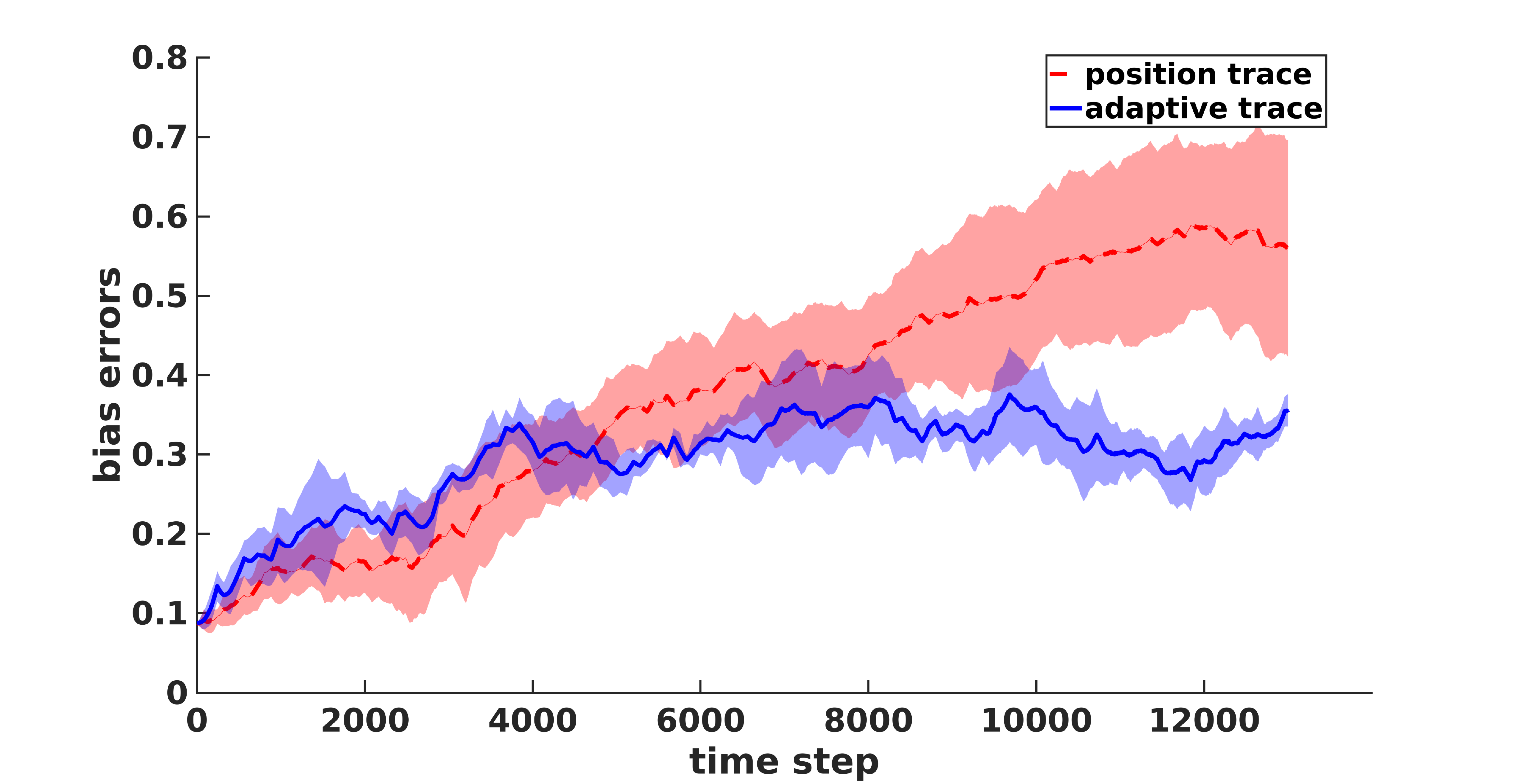}
\caption{ The plots above show bias error over a 600s trajectory, for 50 Monte-Carlo simulation for the adaptive trace and position trace method.}
\label{fig:bias_error_bounds}
\vspace{-1em}%
\end{figure}

This approach is compared against a method which uses the trace of the robot position estimate covariance to guide the planner. A sampling rate of 20Hz is used for sampling the GP regression trajectory. We conduct a 50-run Monte Carlo simulation. Note, for all the experiments, an identical prior trajectory is used to explore the environment first to generate a map of the environment and get a prior estimate for the state covariance. All the experiments are run for the same number of time steps, 12000 which results in 600s trajectories. The mean and standard deviation for the localization and bias RMSE for each of the two approaches are shown in Fig.~\ref{fig:robot_pose_error_bounds}  and Fig.~\ref{fig:bias_error_bounds} respectively.

Initially, the localization errors of the two approaches are comparable, as can be seen in the first 2000 time steps from Fig.~\ref{fig:robot_pose_error_bounds}. However, with more time steps, the localization error of the adaptive traces method is less than that of the method using robot position traces alone. This is because the adaptive method prioritizes convergence of the bias estimates, and better quality bias estimates ultimately lead to improved estimation of the entire state. After the 12000 steps, the average localization error is ${9.846}\,\mathrm{m}$ using the adaptive trace method and ${28.04}\,\mathrm{m}$ with the robot position trace.  We also note how the bias convergence occurs more quicker in the approach where we inform the planner with waypoints that prioritize bias estimate convergence.

\subsection {RRT$^{*}$ optimal path vs greedy planning } 
We compare the performance of our RRT* variant, that uses the adaptive trace as the cost function as it grows the tree vs the greedy algorithm which only picks the best waypoint locally. The RRT$^{*}$ algorithm is limited to 3000 nodes and it is grown without a set goal node as this allows it to be more exploratory. The biases error from the optimal RRT$^{*}$ path are then compared with the average bias errors from the greedy algorithm. 

\begin{figure}[t]
\centering
\includegraphics[width=\linewidth,scale=0.8]{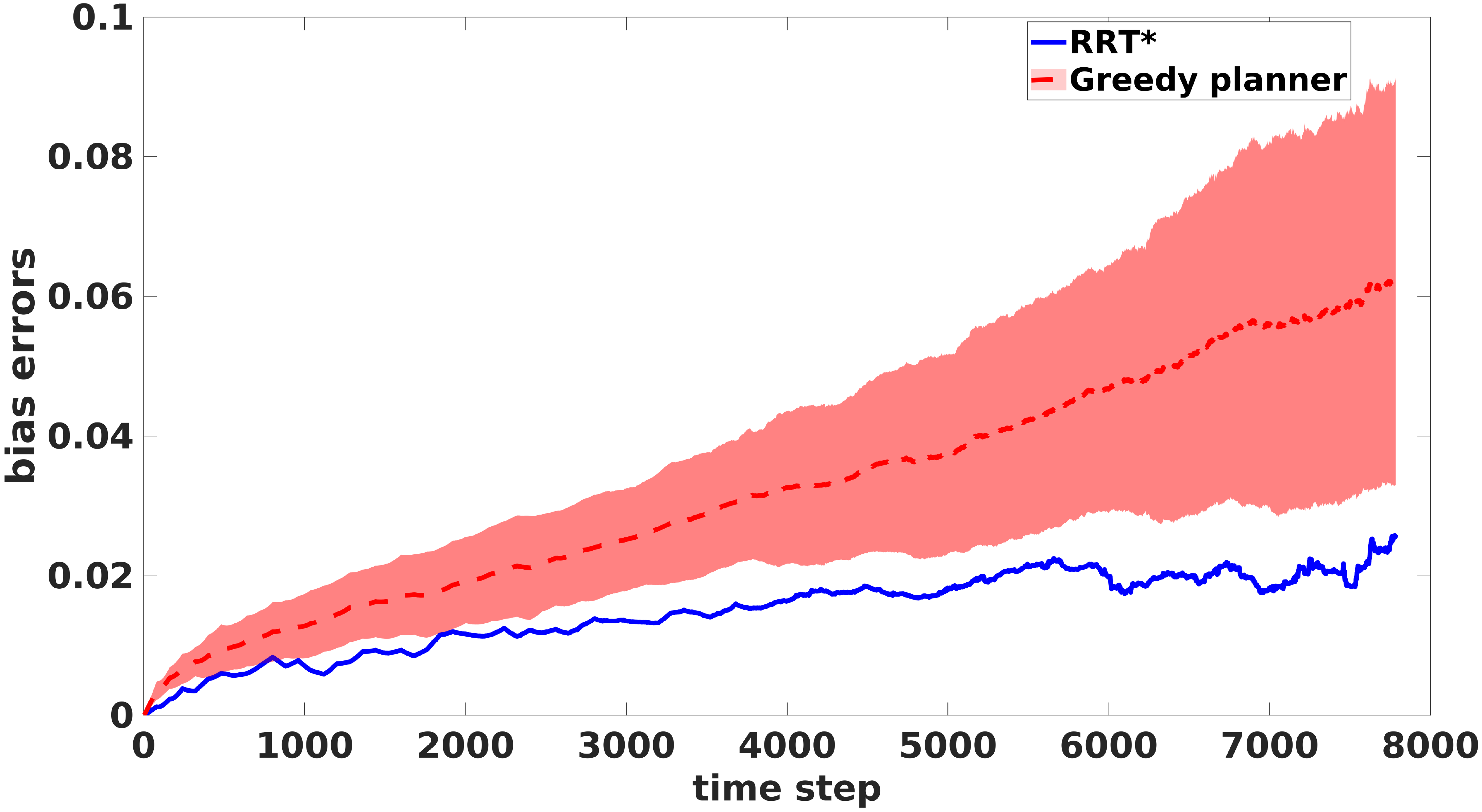}
\caption{Comparing of how bias errors vary over time for the greedy planner and RRT$^{*}$. Both planners use the adaptive trace technique and they are run for 390s.}
\label{fig:rrt vs_greedy_bias_error}
\end{figure}

The results from these simulation are shown in Fig.~\ref{fig:rrt vs_greedy_bias_error}. The plot shows that the bias errors from the RRT$^{*}$ are lower that those for the greedy planner although both planners are using the adaptive trace cost function. At the end of the 390s trajectory, the bias error for the greedy planner is $0.062$ while that for the RRT$^{*}$ planner is $0.026$. This result is consistent with what is expected because the RRT$^{*}$ algorithm has a rewiring technique which ensures that the newest sample is connected on a minimal cost branch to the start node unlike the greedy approach which only considers the cost of connecting the new sample to current node.

\subsection{ GP regression trajectories vs minimal snap trajectory}

We compare the bias and localization error from the GP regression and minimum snap trajectories. In both simulations, 50 Monte-Carlo runs are considered over 300s trajectories and the average errors are compared.

\begin{figure*}
     \centering
     \subfigure[Acceleration on $x$ axis]{
         \includegraphics[width=0.65\columnwidth]{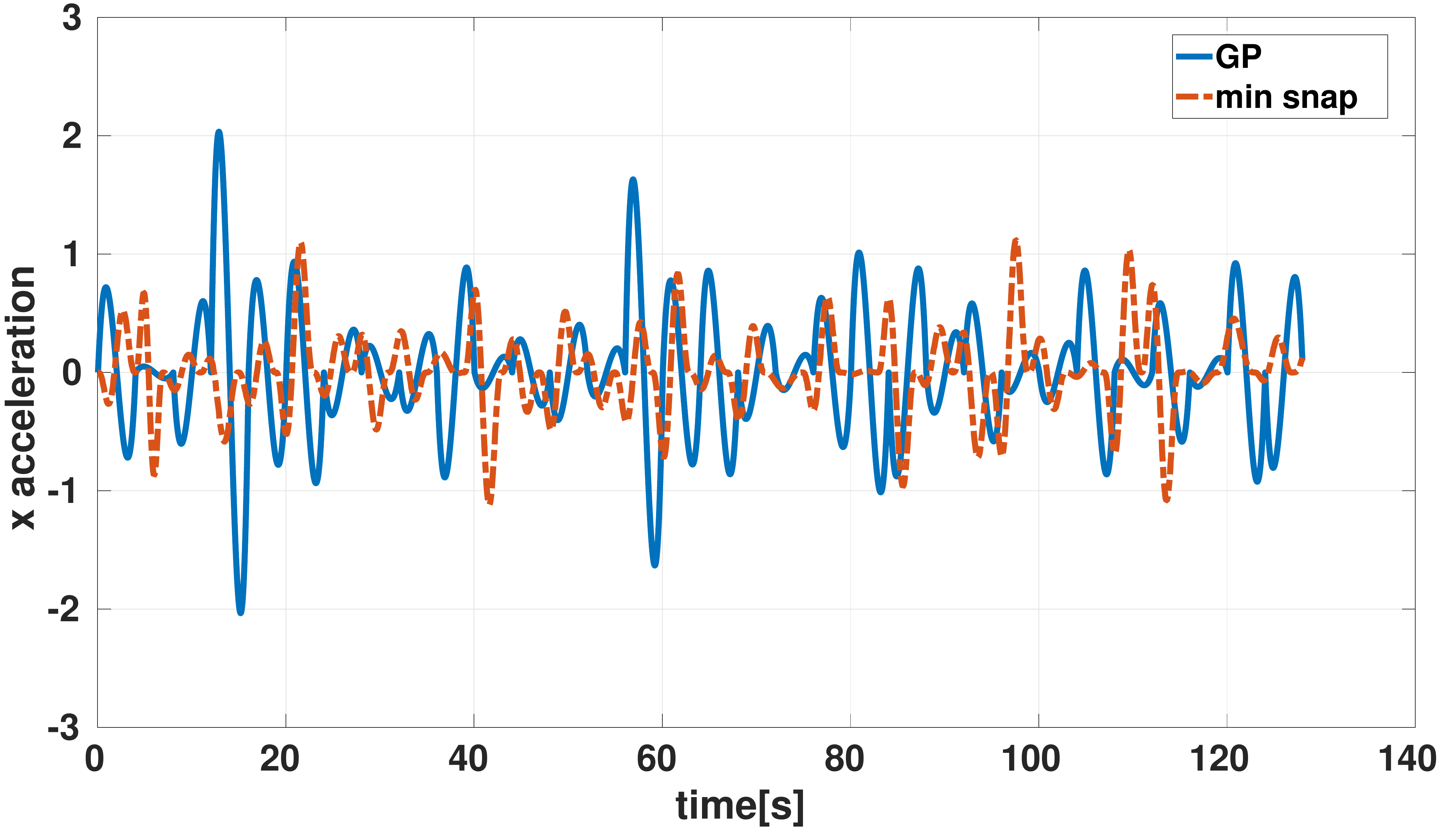}}
     \subfigure[Acceleration on $y$ axis]{
         \includegraphics[width=0.65\columnwidth]{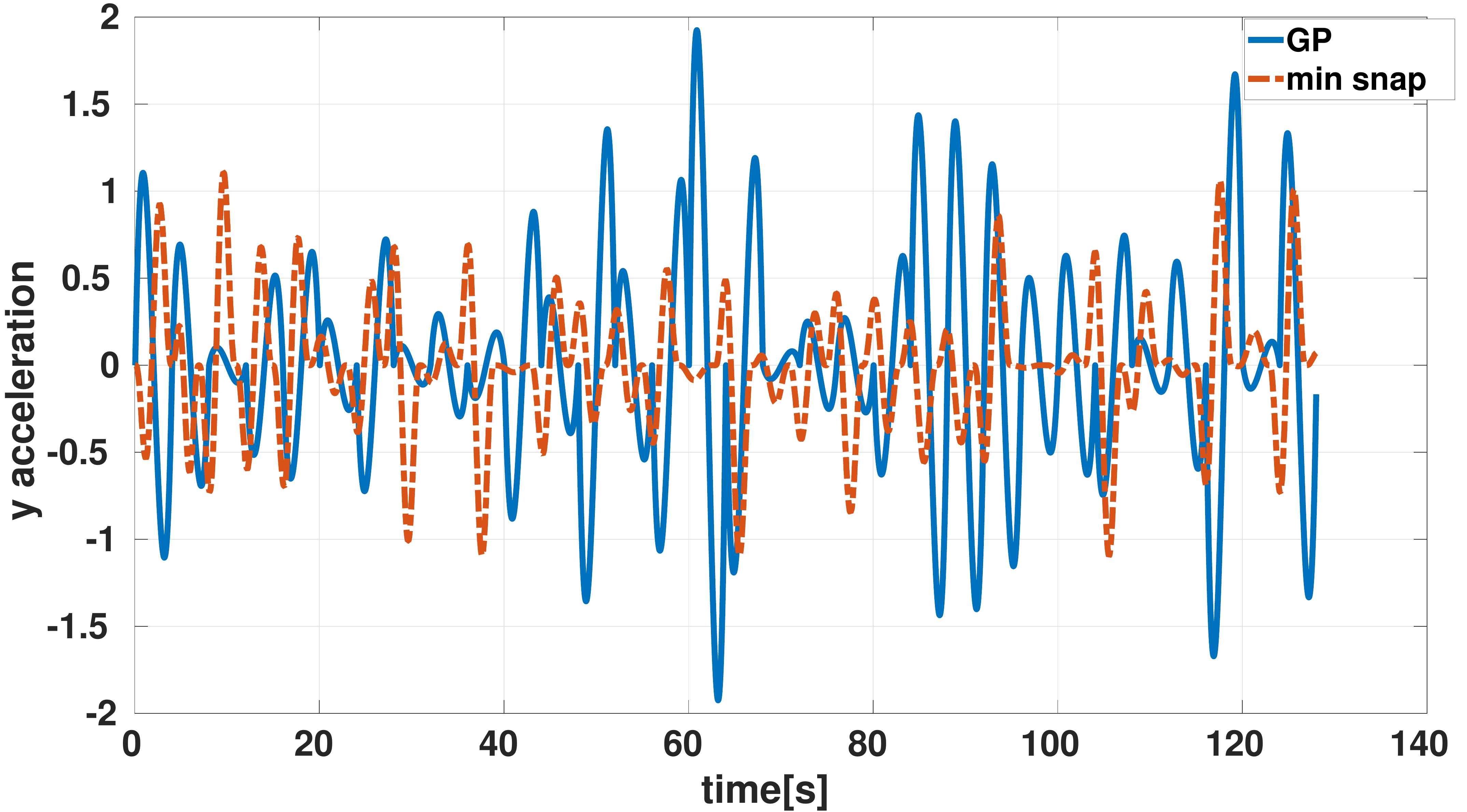}}
     \subfigure[Acceleration on $z$ axis]{
         \includegraphics[width=0.65\columnwidth]{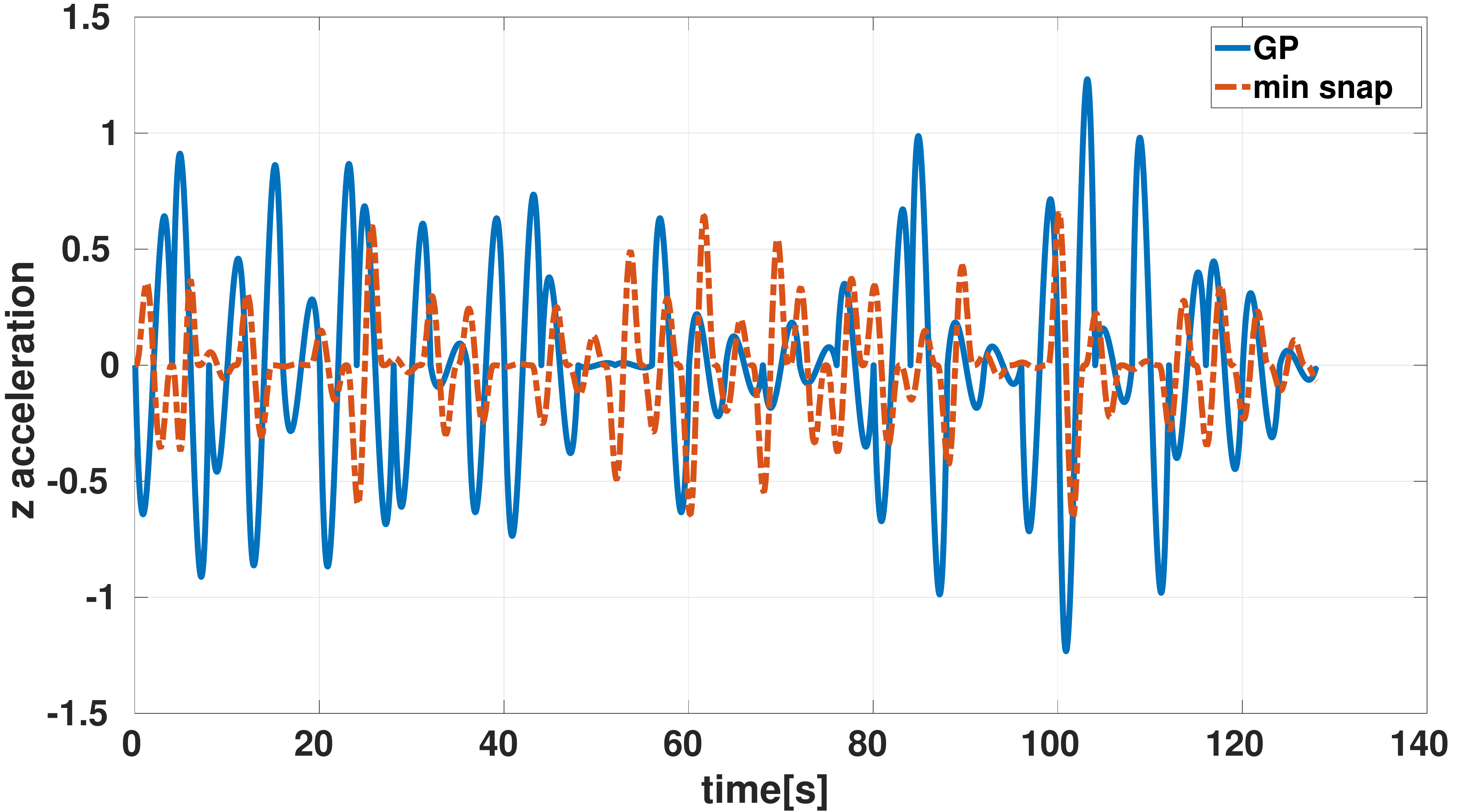}}
        \caption{Accelerations on $x$, $y$ and $z$. The blue trajectory is the GP interpolation acceleration trajectory and the orange trajectory is the minimum snap acceleration trajectory.}
        \label{fig:acc_3_axes_comparison}
\end{figure*}

The results in Table.~\ref{table:1} show that GP interpolation performs better than minimum snap trajectories. We believe that this is the case because the acceleration trajectories from GP regression have larger magnitudes in comparison to those from minimum snap for the same maximum acceleration setting as can be seen in Fig.~\ref{fig:acc_3_axes_comparison} and this generates more excitation for the GP trajectories which leads to quicker convergence of the IMU bias. This ultimately leads to smaller localization errors for GP regression trajectories.

\begin{table}
\begin{center}
\caption{Comparison of the bias and robot position errors accumulated after a 300s trajectory. The first row shows averaged results over 50 Monte-Carlo runs for the Gaussian Process regression interpolation. The second row has averaged results over a 50 Monte Carlo run using minimum snap interpolation method.}
\begin{tabular}{c|c|c}
    \hline \hline
    Interpolation & Average bias & Average localization\\method& RMSE $[m/s^2]$ &  RMSE $[m]$ \\ 
    \hline
    \textbf{GP regression} & \textbf{0.033} & \textbf{2.623}  \\
    minimum snap  & 0.0468    & 5.3913   \\
    \hline
\end{tabular}

\label{table1}
\label{table:1}
\end{center}
\vspace{-2em}
\end{table}

\subsection{Hardware experiment}
In the hardware experiment, we compare the performance of our proposed adaptive trace algorithm using GP regression on a UR5 arm with a stereo camera and an IMU attached. The aim of this experiment is to show the performance in localization of the proposed method with respect to a non-adaptive method using minimum snap trajectories. 

The camera used in this experiment is the Realsense D455 with its internal Bosch BMI055 IMU. The camera provides global shutter RGB images at 20Hz and IMU measurements at 200Hz. The camera and the IMU are calibrated using Kalibr \cite{furgale2013unified}. The state, feature map and the associated covariance matrix are estimated by a formulation of the Iterated Extended Kalman Filter implemented in ROVIO \cite{bloesch2015robust} after execution of our trajectories on the arm. 

Evaluation of the information content in the trajectories generated by our planner is carried out in simulation where we simulate measurements for each of the candidate trajectories, which are evaluated by using the map of the environment and the robot state. The simulated measurements are used to propagate the filter state and its covariance. The planner then decides on the best path to execute by evaluating and comparing the information content in each of the candidate trajectories (see Fig.~\ref{fig:fig_roadmap_robot_features_trajectory}). 

\subsubsection{Robot arm planner}

\begin{figure}[t]
\centering
\includegraphics[ width=\linewidth]{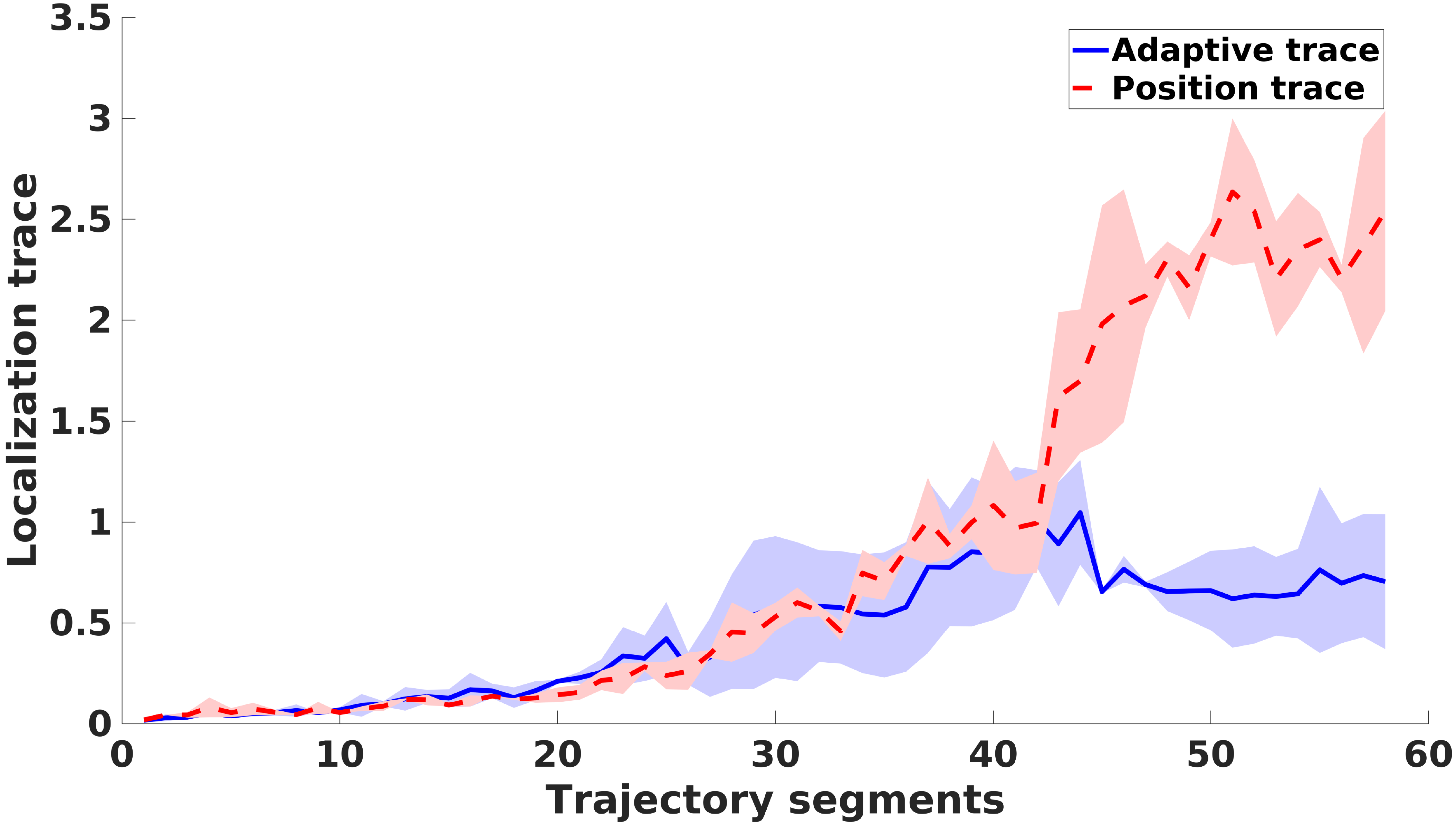}
\caption{Comparing of how localization trace varies with the number of segments added to the trajectory.}
\label{fig:localization_trace}
\vspace{-1em}%
\end{figure}

The planning method proposed in Sec.~\ref{sub_sec:planning} allows for unconstrained sampling of trajectories in $SE(3)$. For the hardware experiments we constrain the trajectories to be executable by the robot arm. Specifically, for a sampled trajectory we require that a valid inverse kinematics (IK) solution exists for each pose, the joint limits of the robot are not violated and the robot does not collide with itself or the environment. Furthermore, we want to avoid any large changes in the arm's configuration between two consecutive poses in order to ensure smooth trajectories which improves tracking and avoids damaging the camera or its cable routed along the robot.

While this could be achieved by sampling directly in the configuration space of the robot, it is computationally expensive and not obvious how to bias sampling in order to achieve diverse excitation for the sensor system. Hence, to enable fast and direct sampling of trajectories in $SE(3)$ we leverage Hausdorff approximation planner (HAP)~\cite{sukkar2022motion} 

which, given a robot, task-space and environment model, computes a subspace in $SE(3)$ to sample from such that the resulting executed robot trajectory satisfies our desired constraints. This subspace is represented using a discrete roadmap of poses, shown in Fig.~\ref{fig:fig_roadmap_robot_features_trajectory}(a), such that moving along a path between any two poses in $SE(3)$ within the subspace results in a similar length path in configuration space.

This roadmap is provided to the RRT$^{*}$ planner to bias its sampling towards. The sampled trajectories are post processed and ensured to be within the provided subspace by checking for time-continuous safety and any large changes in arm configuration between two consecutive poses. If either of these conditions are violated the trajectory is discarded. In practice it was found that a majority of trajectories were within the subspace and not discarded owing to the robustness of the planner.

\subsubsection{Results}
The results of the experiment are shown in Fig. \ref{fig:localization_trace}.
Between segments 0 to 43, the localization traces between the two methods are comparable. However, after convergence of bias uncertainty, the growth of the localization uncertainty is significantly smaller in the experiment using adaptive trace and GP interpolation method as opposed to the non-adaptive method in the dashed red plot. After 58 segments, the localization trace is $0.704$ in the adaptive experiment and $2.540$ in the non-adaptive experiment. This shows that planning for IMU bias convergence helps minimize localization error in state estimation. 

\section{Conclusion}
This paper proposed a new algorithm for informative path planning over continuous trajectories to minimize localization error. The key contribution is the use of Gaussian Process regression to interpolate waypoints coming from our sampling based planner. Linear operators are applied to the kernel function of underlying position GP in order to infer the first and second derivative which are the velocity and acceleration respectively. The use of linear functionals enable velocity and acceleration constraints to be added in the GP model as part of the measurement vector. Furthermore, we proposed an adaptive cost function that used either the robot position trace or the biases trace within the planner in order to prioritize convergence of the IMU biases. This adaptive trace technique is used as the cost function in the RRT$^{*}$ variant that generates a set of discreet waypoints.
Our method is evaluated in 3 simulation experiments and one real world experiment. Overall, our work has shown that planning for IMU bias convergence helps minimize localization error in state estimation.

\bibliographystyle{IEEEtran}
\bibliography{library.bib}
\end{document}